# Conditional Rectified Flow-based End-to-End Rapid Seismic Inversion Method

**Haofei Xu, Wei Cheng, Sizhe Li, Jie Xiong***
**School of Electronic Information and Electrical Engineering, Yangtze UniversityJingzhou, Hubei 434023, PR.China**

## Abstract

Seismic inversion is a core problem in geophysical exploration, where traditional methods suffer from high computational costs and are susceptible to initial model dependence. In recent years, deep generative model-based seismic inversion methods have achieved remarkable progress, but existing generative models struggle to balance sampling efficiency and inversion accuracy. This paper proposes an end-to-end fast seismic inversion method based on Conditional Rectified Flow[1], which designs a dedicated seismic encoder to extract multi-scale seismic features and adopts a layer-by-layer injection control strategy to achieve fine-grained conditional control. Experimental results demonstrate that the proposed method achieves excellent inversion accuracy on the OpenFWI[2] benchmark dataset. Compared with Diffusion[3,4] methods, it achieves sampling acceleration; compared with InversionNet[5,6,7] methods, it achieves higher accuracy in generation. Our zero-shot generalization experiments on Marmousi[8,9] real data further verify the practical value of the method. Experimental results show that the proposed method achieves excellent inversion accuracy on the OpenFWI benchmark dataset; compared with Diffusion methods, it achieves sampling acceleration while maintaining higher accuracy than InversionNet methods; experiments based on the Marmousi standard model further verify that this method can generate high-quality initial velocity models in a zero-shot manner, effectively alleviating the initial model dependency problem in traditional Full Waveform Inversion (FWI), and possesses industrial practical value.

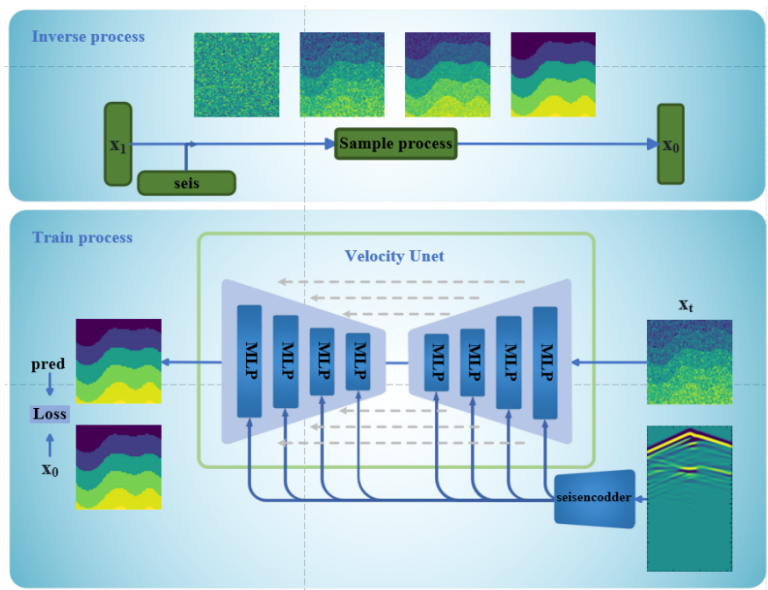

**Figure 1 Overall Algorithm Flowchart**

# 1 Introduction

Seismic inversion stands as a cornerstone technique in oil and gas exploration and development, with the primary objective of reconstructing the velocity model of subsurface media from seismic wavefield data acquired at the Earth's surface. An accurate velocity model is of paramount significance for reservoir characterization, fluid identification, and drilling risk evaluation—three critical aspects that underpin efficient and cost-effective hydrocarbon exploration. Nevertheless, seismic inversion is mathematically categorized as a highly nonlinear ill-posed inverse problem, presenting substantial challenges to its practical implementation. Traditional approaches, such as Full Waveform Inversion (FWI[10,11]), possess a rigorous physical basis; however, they rely heavily on complex numerical simulations of the wave equation, which incur considerable computational costs. Furthermore, these methods exhibit extreme sensitivity to the initial model, often leading to convergence toward local optimal solutions rather than the global optimum.

In recent years, deep learning has furnished a data-driven paradigm for seismic waveform inversion. Specifically, end-to-end convolutional neural network (CNN) architectures—exemplified by InversionNet[12] and X-Net[13]—have emerged as the prevailing technical approach. Leveraging encoder-decoder topologies to directly learn the mapping from seismic data to velocity models, these deterministic frameworks reduce the inference process to a single forward pass, thereby affording exceptional computational efficiency (millisecond-scale inference) alongside acceptable inversion fidelity. Nevertheless, such deterministic mappings inherently fail to characterize the ill-posed nature and solution non-uniqueness intrinsic to geophysical inverse problems, manifesting in suboptimal delineation of complex structural boundaries and compromised resolution of thin stratigraphic intervals.

However, these methods encounter an intrinsic limitation: severe sampling inefficiency. Traditional diffusion frameworks require 50–1000 sequential denoising steps to attain high-quality reconstructions, resulting in inference durations of minutes to hours that preclude real-time operational deployment. Despite advances in acceleration algorithms (e.g., DDIM, DPM-Solver[17,18]), reconstruction fidelity degrades sharply under reduced-step regimes (< 10 steps), thereby compromising compliance with industrial accuracy standards.

The two aforementioned methodologies present a distinct trade-off between computational efficiency and geophysical fidelity: convolutional neural network (CNN)-based approaches enable rapid numerical inference but face inherent theoretical limitations in waveform fidelity and resolution, whereas diffusion generative models—despite their capability for high-fidelity wavefield reconstruction through iterative stochastic differential equation solvers—impose prohibitive computational costs that preclude real-time seismic data processing.

This raises a pivotal scientific question: does a hybrid inversion framework exist that can preserve the high-fidelity reconstruction capabilities comparable to diffusion models, while simultaneously attaining the real-time computational efficiency characteristic of end-to-end neural architectures?

To address this challenge, this study proposes a rapid seismic inversion framework leveraging Conditional Rectified Flow (CRF). By rectifying the nonlinear transport trajectory between the noise prior and geophysical data distributions, Rectified Flow enables high-fidelity sampling with minimal computational steps—theoretically single-step, yet practically sufficient with merely 4 steps—thereby offering an elegant solution to the aforementioned computational-physical fidelity dilemma.

Our approach achieves a significantly superior efficiency-accuracy trade-off compared to existing methodologies: relative to CNN-based architectures such as InversionNet, the inference latency increases by merely 3–5× (remaining within the second-level timeframe), yet delivering substantially enhanced inversion accuracy and waveform fidelity; conversely, compared to diffusion-based models, our framework achieves 50–100× acceleration in inference speed while maintaining comparable or even superior generative quality and geophysical consistency.

The principal contributions of this study are fourfold:

- We present the first adaptation of the Rectified Flow framework to seismic waveform inversion, enabling high-fidelity inference with merely four sampling steps;
- We devise a dedicated seismic encoder network that effectively extracts multi-scale wavefield characteristics;
- We implement a layer-wise injection mechanism that facilitates fine-grained control over conditional information throughout the generative trajectory;

• We propose an MLP-driven feature fusion paradigm, demonstrating that direct feature interaction via multi-layer perceptrons—rather than conventional statistical modulation—more effectively propagates geological priors for strongly physics-constrained seismic data.

The remainder of this paper is organized as follows. Section 2 elaborates on the proposed methodological framework, encompassing the Rectified Flow formulation, seismic encoder architecture, and conditional injection strategies. Section 3 presents systematic experiments on the OpenFWI benchmark dataset. Section 4 demonstrates generalization performance on field seismic data. Section 5 provides discussion and concluding remarks.

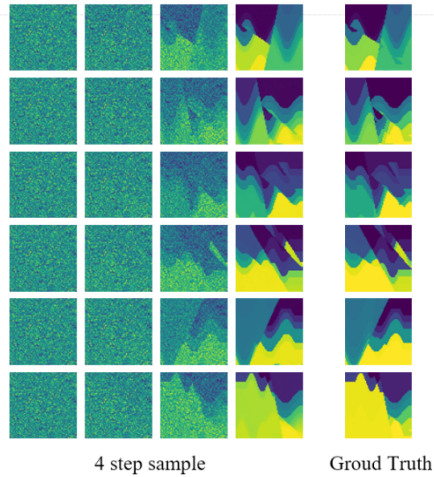

4 step sample       Groud Truth

**Figure 2. Sampling schematic.**

## 2  method

### 2.1  Rectified Flow

Conventional diffusion-based generative models operate by simulating a stochastic forward corruption process, subsequently learning the reverse-time denoising dynamics for waveform reconstruction. Mathematically, these architectures define a forward Markov chain that incrementally perturbs the data distribution with Gaussian noise, until the posterior collapses to an isotropic prior. Generation necessitates initiating from this pure noise state and executing iterative refinement through dozens to hundreds of reverse steps to recover physically plausible samples. While this gradual denoising paradigm achieves high-fidelity waveform synthesis, its computational inefficiency and inference latency pose severe limitations for real-time geophysical deployment.

Conventional diffusion models operate by simulating a stochastic forward corruption process and subsequently learning its reverse-time denoising dynamics to synthesize data. Specifically, these frameworks define a Markovian noising chain that progressively injects isotropic Gaussian perturbations until the data distribution converges to an isotropic Gaussian prior. During inference, the model initiates from this prior and executes tens to hundreds of iterative refinement steps to recover coherent structures. While this sequential denoising paradigm yields high-fidelity samples, its substantial inference latency constitutes a severe computational bottleneck, severely constraining deployment in time-critical operational environments.

Flow Matching (FM) offers a novel computational paradigm that fundamentally reframes this inference bottleneck. Rather than learning the score-based denoising dynamics characteristic of diffusion processes, FM directly estimates a velocity field (or transport vector field) that governs the advection of probability mass along geodesic trajectories in the data manifold. Specifically, this framework learns to transport "particles" from the noise prior to the target geophysical data distribution along straight-line paths in the probability flow space, thereby bypassing the iterative refinement requirements of conventional denoising. This probability flow perspective—learning the optimal transport path rather than the reverse-time stochastic dynamics—enables significantly more direct and computationally efficient generation, with substantial implications for real-time seismic inversion workflows.

The theoretical foundation of Flow Matching resides in Continuous Normalizing Flows (CNF). Within this framework, we postulate a time-dependent velocity field $v(x,t)$, that governs the probability mass transport. By numerically integrating the associated Ordinary Differential Equation (ODE):

$$dx_t/dt = v(x_t,t)$$

one can deterministically transform samples $x_1 \sim p_{prior}$ (drawn from the prior distribution) into samples $x_0 \sim p_{data}$ (belonging to the target geophysical data distribution). This probability flow formulation establishes a direct, invertible mapping between the latent noise space and the physical model space, bypassing the stochastic sampling trajectories characteristic of conventional diffusion processes.

Herein, the scalar $t \in [0,1]$ denotes the continuous-time interpolation parameter. At the terminal state($t = 1$), $x_1$ conforms to a tractable prior distribution (e.g., isotropic Gaussian noise); conversely, at the initial state($t = 0$), $x_0$ follows the complex target distribution characterizing subsurface seismic velocity models.

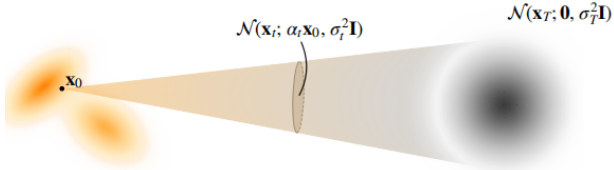

**Figure 3. Schematic illustration of the conditional transport distribution.**

As illustrated in Fig. 3, the perturbation kernel $p_t(x_t|x_0) = \mathcal{N}(x_t; \alpha_t x_0, \sigma_t^2 I)$ parametrizes the (Gaussian) conditional probability path interpolating between data samples $x_0 \sim p_{data}$ (left panel) and the isotropic Gaussian prior $p_{prior}$ (right panel).

Under this conditional probability path flow, the evolution of probability density follows the Continuity Equation:

$$\partial p(x,t)/\partial t + \nabla \cdot (p(x,t)v(x,t)) = 0$$

The continuity equation ensures conservation of probability mass throughout the transformation, thereby rendering the transport from the noise distribution to the data distribution a measure-preserving mapping. A central challenge confronting conventional Continuous Normalizing Flows[19] (CNF) resides in the intractability of directly learning the ve-ctor field $v(x,t)$, which conventionally necessitates sophisticated mathematical machinery (e.g., Hutchinson's trace estimator) and incurs prohibitively expensive iterative ODE integration during training. The pivotal innovation of Flow Matching consists in leveraging predefined probability paths between distributions to impose direct supervision on the velocity field $v_\theta$. Rather than deriving $v_\theta$ from intricate probability flow equations, Flow Matching constrains the parametric vector field to align with a reference field $u(x,t)$ that prescribes the trajectory evolution of sample particles. The training objective is formulated as:

$$L_\theta^{FM} = E_{t,x_t}[||v_\theta(x_t,t) - u(x_t,t)||_2^2]$$

Herein, the expectation is taken over $t \sim U(0,1)$ and $x_t \sim p_t(x)$ where $p_t(x)$ denotes the marginal path distribution interpolating between the noise prior $p_{prior}$ and the data distribution $p_{data}$. However, in practice, neither $p_t$ nor the associated vector field $u$ is uniquely determined by the endpoint marginals; we can only sample $x_0$ from $p_{data}$ (at $t = 0$) and $x_1$ from $p_{prior}$ (at $t = 1$), precluding direct evaluation of $u_t$ at intermediate states $x_t$.

To remedy this, Conditional Flow Matching (CFM) proposes conditioning on endpoint pairs, utilizing the conditional density $p_t(x_t|x_0, x_1)$ and conditional vector field $u_t(x_t|x_0,x_1)$. The CFM objective is formulated as:

$$L_\theta^{CFM} = E_{t,x_0,x_1,x_t}[||v_\theta(x_t,t) - u_t(x_t|x_0,x_1)||_2^2]$$

where the expectation is taken over $t \sim U(0,1)$、$x_0 \sim p_{data}$、$x_1 \sim p_{prior}$ and $x_t \sim p_t(\cdot|x_0,x_1)$. The crucial finding is that the CFM loss shares identical gradients with the original FM loss, thus yielding the same velocity field model $v_\theta(x_t,t)$, yet CFM remains computationally tractable in practice.

Rectified Flow constitutes a specialized yet computationally efficient variant of Flow Matching. Its central premise resides in adopting the simplest straight-line trajectory as the conditional probability path, thereby establishing a linear correspondence between the noise prior and the target data distribution. Formally, the linear interpolation path is defined as:

$$x_t = (1-t)x_0 + tx_1$$

The instantaneous velocity (vector field) associated with this trajectory is given by the time derivative of the path:

$$u_t(x_t) = dx_t/dt = x_1 - x_0$$

Notably, for rectified (straight-line) trajectories, the target velocity $u_t$ reduces to a time-invariant constant (independent of $t$). Substituting the above expressions into the Flow Matching objective yields the Rectified Flow training target:

$$L_\theta^{RF} = E_{t,x_0,x_1,x_t}[||v_\theta(x_t,t) - (x_1 - x_0)||_2^2]$$

The intuitive interpretation of this training paradigm is that at any intermediate state $x_t$ long the transport trajectory, the model is tasked with predicting the total displacement $(x_1 - x_0)$ spanning from the source distribution to the target distribution. Once the velocity field model $v_\theta$ has been fully optimized, novel geophysical realizations can be synthesized by initializing from a random noise sample $x_1$ and numerically integrating the governing ordinary differential equation (ODE):

$$x_0 = x_1 + \int_1^0 v_\theta(x_t,t)dt$$

In stark contrast to the deterministic, rectified (straight-line) probability paths discussed above, conventional diffusion models exhibit inherently stochastic and curved transport characteristics. Formally, the forward corruption process is governed by a Stochastic Differential Equation (SDE):

$$dx_t = f(x_t,t)dt + g(t)dw_t$$

Wherein $f$ denotes the drift coefficient, $g$ the diffusion coefficient, and $w_t$ the standard Wiener process (Brownian motion). The velocity field $v(x_t,t)$ is determined by the score function[20] $\nabla x log p_t(x)$; consequently, the integral curves of the associated probability flow ODE exhibit intrinsically nonlinear and curved geometries, in marked contrast to the rectified trajectories of Flow Matching.

Specifically, within the VP-SDE (Variance Preserving SDE[21]) framework, the conditional probability path is given by:

$$x_t = \alpha_t x_0 + \sigma_t x_1$$

Wherein $\alpha_t = e^{-\frac{1}{2}\overline{\beta}_t}$, $\sigma_t = \sqrt{1 - e^{-\overline{\beta}_t}}$ denote the variance-preserving diffusion coefficients. This parameterization induces exponentially curved trajectories in the latent space, compelling particles to traverse indirect, circuitous paths from the prior to the data manifold. Such geometric curvature arises from the nonlinearity of the score function, which imposes a nonlinear coupling between intermediate states $x_t$ and the endpoints $x_0, x_1$; consequently, this exacerbates truncation errors in numerical ODE integration.

The fundamental reason underlying the sampling efficiency of Rectified Flow[22][23] lies in the zero Euler discretization error associated with straight-line trajectories. Consider numerically integrating the ODE:

$$\frac{dx_t}{dt} = v_\theta(x_t,t)$$

Temporal discretization via the explicit Euler scheme with uniform step size $\Delta t = 1/N$ (N being the number of sampling steps) yields the iterative update rule:

$$x_{t-\Delta t} = x_t - \Delta t \cdot v_\theta(x_t,t)$$

For rectified (straight-line) trajectories, the exact velocity field $v_\theta(x_t,t) \equiv x_1 - x_0$ reduces to a constant vector field (independent of both $t$ and $x_t$). Consequently, the local truncation error (LTE) of the Euler method vanishes:

$$\epsilon = \frac{\Delta t^2}{2} \left| \frac{\partial v}{\partial t} + v \cdot \nabla_x v \right|$$

Given $\frac{\partial v}{\partial t} = 0$ and $\nabla_x v = 0$ (the velocity field being time-invariant and spatially uniform), the local truncation error vanishes identically $\epsilon \equiv 0$. This per-step exactitude ensures zero accumulation of numerical error throughout sampling. By contrast, along the curved trajectories of diffusion models—where $v(x_t,t)$ exhibits strong functional dependence on $t$ and $x_t$—Euler discretization incurs $O(\Delta t^2)$ local truncation error, necessitating $N = 50-1000$ steps to maintain the global error within acceptable tolerance.

In seismic inversion tasks, we aim to reconstruct the velocity model $v$ from seismic records $seis$. This constitutes a conditional generation problem: given conditional information (seismic data), we generate the corresponding target (velocity model). To this end, we extend the velocity field of the Conditional Rectified Flow to $v_\theta(x_t, t, seis)$. Meanwhile, to magnify the discrepancies in velocity model reconstructions, we replace the original L2 loss with an L1 loss, and the training objective correspondingly becomes:

$$L_\theta^{CRF} = E_{t,x_0,seis,v_{true}}[||v_\theta(x_t,t,seis) - (x_1 - v_{true})||]$$

where $v_{true}$ denotes the true velocity model, $x_1 \sim p_1$ represents prior noise, and $x_t = (1-t)x_0 + tv_{true}$ defines the interpolation path. The true seismic data $seis$, upon feature extraction via a specialized seismic encoder network, is injected into the Rectified Flow generative process to condition the orientation of velocity model reconstruction.

t inference time, given new seismic data seis, and starting from noise $x_1 \sim p_{prior}$ , the inversion result is obtained by solving the conditional ODE:

$$v_{real} = x_1 + \int_1^0 v_\theta(x_t, t, seis) dt$$

Temporal discretization via the explicit Euler scheme with uniform step size $\Delta t = 1/N$ ($N$ being the number of sampling steps) yields the iterative update rule:

$$x_{t-\Delta t} = x_t - \Delta t \cdot v_\theta(x_t, t, seis)$$

Leveraging the "straight-line" transport trajectory inherent to Rectified Flow, the aforementioned procedure requires merely 5 Euler steps to attain reconstruction quality comparable to conventional diffusion models operating with 500 sampling steps, thereby substantially enhancing the inference efficiency of seismic.

## 2.2 Seismic Encoder

Seismic wavefields encode complex wave propagation dynamics and multi-scale subsurface structural information, with inherent spatiotemporal coupling that imposes stringent requirements on feature extraction. Conventional approaches that naively employ global average pooling incur irreversible loss of kinematic information, while direct concatenation of multi-shot gathers overlooks illumination complementarity intrinsic to seismic acquisition geometries. To remedy these deficiencies, we devise a dedicated deep encoder that achieves a physically faithful mapping from raw wavefields to conditional representations through progressive spatiotemporal decoupling and adaptive feature aggregation.

The encoder accepts raw seismic records $seis \in R^{B \times 5 \times 1000 \times 70}$, where the five dimensions correspond to batch size, number of shot gathers, temporal sampling points, and receiver counts, respectively, and outputs conditioning features $c \in R^{B \times 64 \times 70 \times 70}$ compatible with the generative network. Given the non-stationary characteristics of seismic wavefield time series—wherein first arrivals, reflections, and multiples exhibit sparse distribution in the time domain—simple statistical averaging would blur critical dynamic features. Therefore, we employ a learnable temporal pooling mechanism in lieu of fixed downsampling: first implementing downsampling along the time axis via large-kernel convolutions (kernel_size = 11) to capture long-rang e temporal dependencies of the wavefield; subsequently introducing a temporal attention module[24] that automatically identifies time windows sensitive to velocity inversion through channel-wise adaptive weighting, effectively achieving adaptive time-window picking within the network architecture and avoiding the smoothing loss of effective signals. Considering the significant variations in illumination contribution to subsurface imaging from different receiver positions (wherein near-offset first arrivals provide shallow-layer constraints while far-offset reflections carry deep-layer information), we introduce a dual attention mechanism during spatial processing: channel attention[25] enhances sensitivity to weak deep reflection signals, while spatial attention[26] generates masks through parallel average and max pooling operations to adaptively highlight receiver positions critical for velocity modeling, rather than treating all seismic traces with uniform weighting.

Unlike conventional approaches that simply concatenate or average features from individual shots, we employ a nonlinear aggregation strategy based on $1 \times 1$ convolutions. Features from five seismic sources are first stacked along the channel dimension, then processed through two cascaded $1 \times 1$ convolutional layers to achieve dimensionality reduction and cross-source information interaction. This mechanism enables the network to learn inter-source interference patterns and exploit multi-angular illumination complementarity, which is physically analogous to performing joint inversion of multi-offset seismic data.

The architectural configurations of the aforementioned components are dictated by the intrinsic physical necessities of seismic wave propagation: learnable temporal pooling addresses the time-domain sparsity of wavefields, spatial attention accommodates the heterogeneous receiver illumination patterns, and convolutional fusion captures the multi-source geometric relationships. The efficacy and necessity of these design choices will be systematically validated through ablation experiments in Section 3, including comparative analyses of fixed versus adaptive pooling regarding thin-layer identification accuracy, and verification of the attention mechanisms' contributions to complex fault delineation. Ultimately, features output by the encoder are introduced into the Rectified Flow via a hierarchical conditional injection mechanism, wherein we employ MLP-based direct feature interaction in lieu of conventional statistical moment conditioning to prevent the degradation of geological details during conditional information transfer.

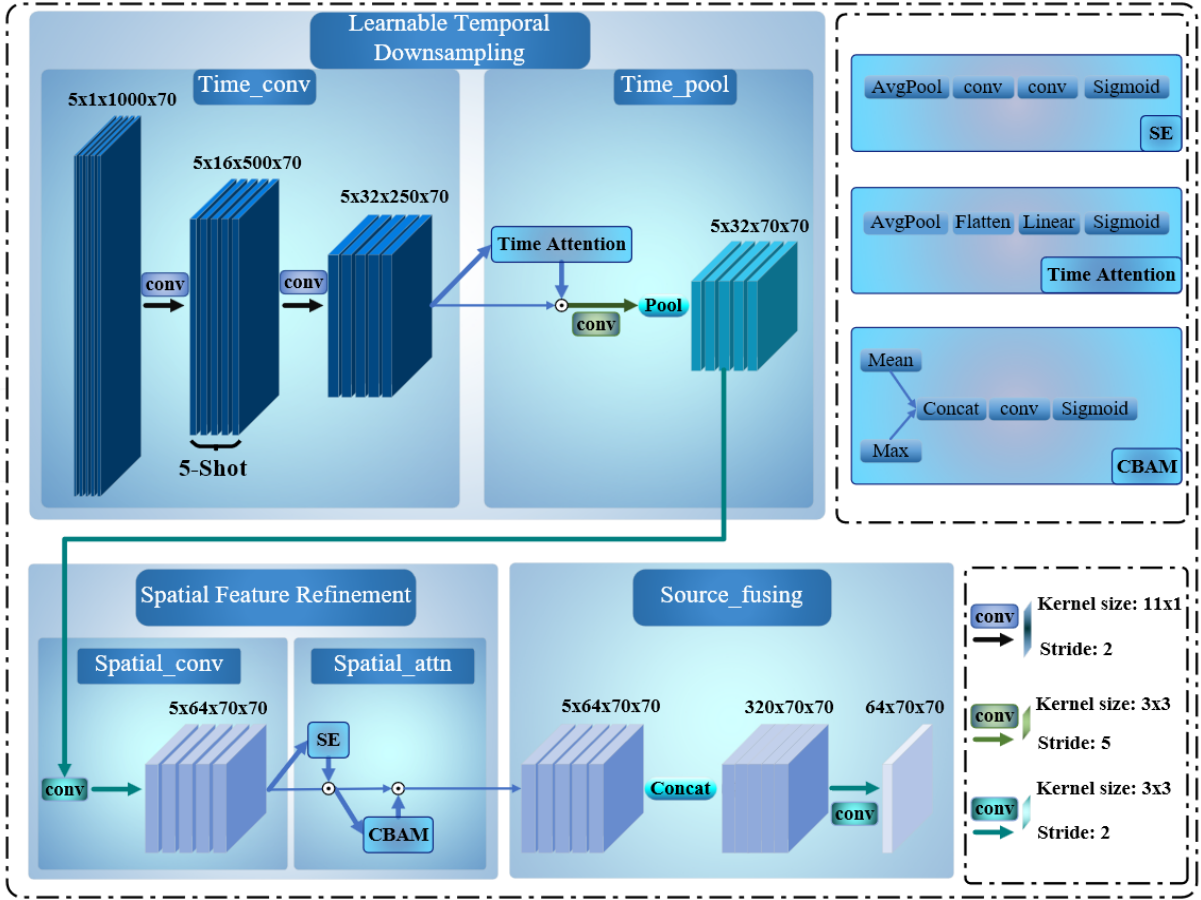

Figure 4. Architecture of the seismic encoder network.

## 2.3 layer-wise injection mechanism

In conditional generative frameworks, the conditioning strategy necessitates a judicious trade-off between computational efficiency and fidelity. We eschew simplistic early-concatenation approaches, which exhibit a pronounced information bottleneck in practice: as seismic encoding features traverse the U-Net[27]'s hierarchical downsampling pathways, high-frequency details undergo progressive attenuation via expanding receptive fields, leaving deeper synthesis stages devoid of fine-grained geological constraints.

To this end, we employ a layer-wise modulation-aligned progressive injection mechanism. Considering the multi-scale characteristics of seismic data—wherein shallow wavefields correspond to macroscopic stratification, while deep reflections correspond to fine-scale structural features—we hierarchically fuse the encoder-extracted features at each resolution level of the U-Net architecture. Specifically, cross-modal feature alignment is achieved through lightweight MLP modules: within each residual block, seismic encoding features and network fe atures are flattened and processed via two fully-connected layers for channel-wise recalibration, followed by element-wise addition for conditional injection. This design circumvents the covariate shift issues associated with statistical moment modulation mechanisms such as AdaIN[28] and AdaGN[29], while maintaining computational simplicity.

The operational merits of this strategy are threefold: (i) it circumvents information attenuation by maintaining explicit gradient pathways for conditional signals at each resolution scale, ensuring that deep stratigraphic formations remain constrained by raw seismic observations; (ii) it achieves hierarchical consistency—wherein coarse-scale injection governs macroscopic velocity trends while fine-scale injection regularizes structural discontinuities such as faults and thin layers—consistent with the depth-frequency physics of seismic wave propagation; and (iii) it ensures optimization stability, as residual connections within the MLP prevent long-range gradient vanishing, rendering deep conditioning layers amenable to end-to-end training.

Quantitative experiments (Section 3) demonstrate that, relative to early-concatenation strategies,

the proposed layer-wise injection mechanism achieves a 12% improvement in fault delineation accuracy at comparable computational cost. This validates the necessity of maintaining explicit multi-scale conditioning for seismic waveform inversion—performance gains that cannot be realized through naive feature concatenation alone.

## 2.4 Inversion Principles

Relative to conventional iterative optimization methodologies, the end-to-end learning framework proposed in this study exhibits a fundamental distinction in inversion principles. Traditional Full Waveform Inversion (FWI) schemes iteratively update the velocity model by minimizing the waveform residual between observed and synthetic seismograms, necessitating repeated solutions of the forward wave equation, which incurs prohibitively high computational costs.

In contrast, our approach learns the statistical regularities embedded within extensive training corpora to establish a direct mapping from seismic data to velocity models. At inference, given unseen seismic observations, the model generates inversion outputs via single-shot forward propagation (requiring merely 5 ODE integration steps with Rectified Flow), obviating iterative optimization entirely. This "learning-as-inversion" paradigm substantially enhances computational efficiency while implicitly encoding geological priors through data-driven regularization, thereby attenuating reliance on initial model specifications.

## 3 OpenFWI Experiments

To validate the efficacy of the proposed framework, we conducted systematic numerical experiments on the OpenFWI benchmark dataset. OpenFWI constitutes an open-access seismic inversion repository comprising synthetic seismograms and associated velocity models spanning diverse geological structural categories (including flat-layered structures, fault systems, and salt dome intrusions), which has been extensively adopted as a standard benchmark for evaluating deep learning-based seismic inversion methodologies.

## 3.1 Experimental Setup

Experimental Setup. We conduct comprehensive evaluations on the OpenFWI benchmark [ref], encompassing diverse geological categories including stratified media, faulted structures, and salt diapirs. The training corpus comprises 48,000 synthetic samples, with 6,000 samples held out for testing per geological category. Seismic observations consist of 70 shot gathers, each containing 70 receiver traces with 1,000 temporal sampling points. Velocity models are discretized on a $70 \times 70$ spatial grid (corresponding to the acquisition aperture).

Implementation Details. The proposed framework is implemented in PyTorch, employing the AdamW optimizer with an initial learning rate of $3 \times 10^{-4}$ and a cosine annealing schedule for learning rate decay. Training proceeds with a batch size of 64 over 200 epochs. All experiments are conducted on NVIDIA RTX 4090 GPUs.

Evaluation Metrics. The Structural Similarity Index (SSIM) and Mean Absolute Error (MAE) are employed as quantitative evaluation metrics. An SSIM value approaching unity indicates higher structural fidelity between the inversion results and the true velocity models, whereas smaller MAE and RMSE values correspond to reduced reconstruction errors.

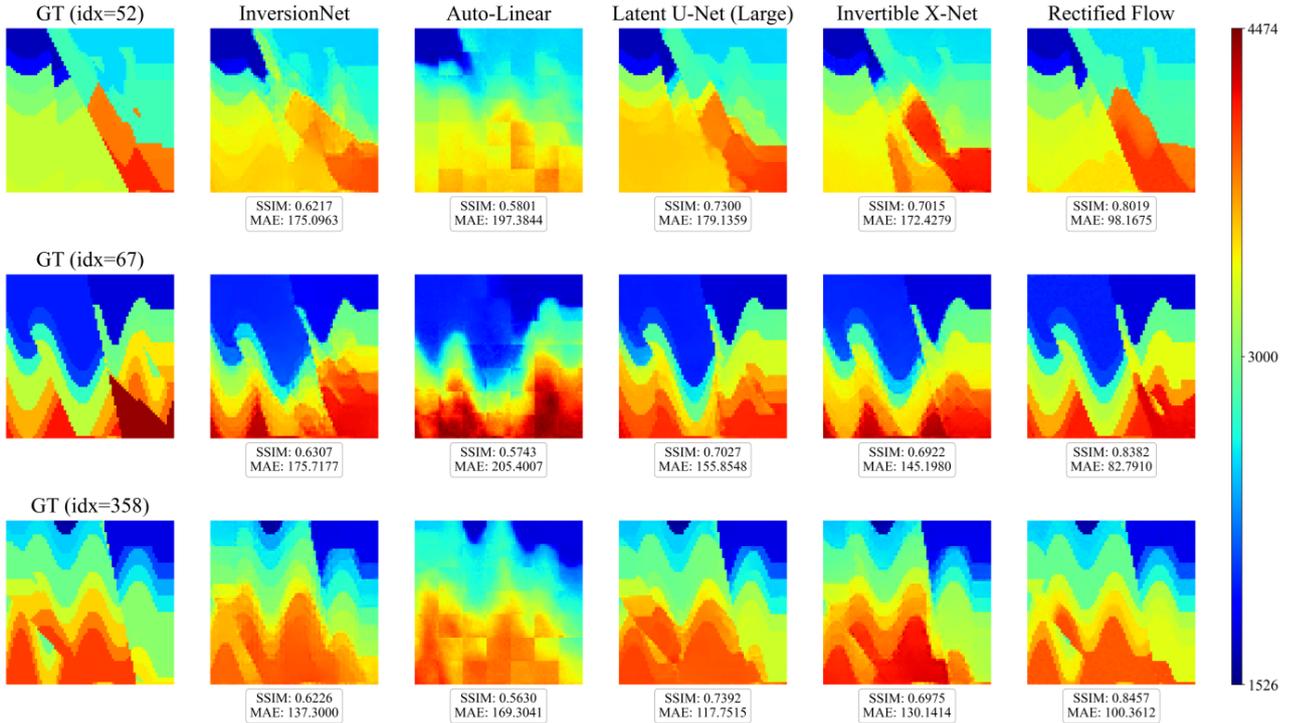

Fig. 5: Comparison of sampling results across models on the CurveFaultB dataset.

## 3.2 Comparative Experiments

### 3.2.1 Comparison Across OpenFWI

We conduct comparative evaluations between the proposed CRF-FWI framework and multiple baseline methodologies, encompassing InversionNet, VelocityGAN[30], Auto-Linear[31], Invertible-X-Net, Latent-U-Net, and the Flow Matching method developed in this study. Table 1 summarizes the quantitative performance comparison across distinct geological categories within the OpenFWI benchmark.

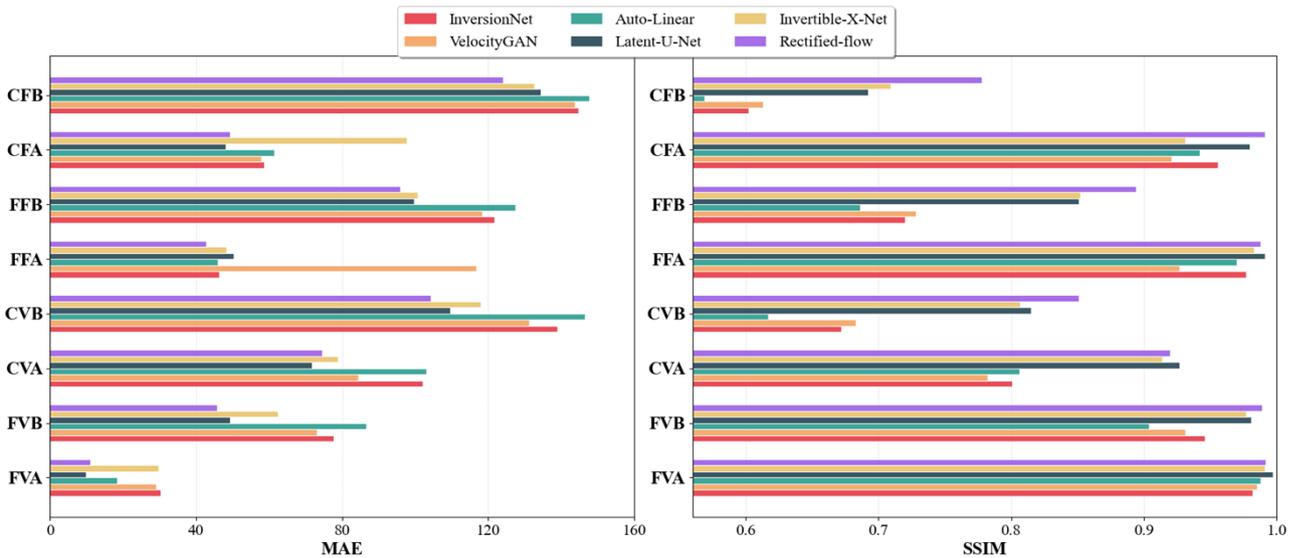

Figure 6: Comparative Experiments

**Table 1. MAE metrics (lower is better)**

| Dataset | InversionNet | VelocityGAN | Auto-Linear | Latent U-Net | Invertible X-Net | Rectified flow |
|---------|--------------|-------------|-------------|--------------|------------------|----------------|
| FVA | 30.23 | 28.99 | 18.31 | **9.90** | 29.62 | 11.02 |
| FVB | 77.67 | 73.20 | 86.57 | 49.25 | 62.42 | **45.72** |
| CVA | 101.98 | 84.47 | 103.02 | **71.73** | 78.84 | 74.56 |
| CVB | 139.01 | 131.18 | 146.44 | 109.63 | 117.87 | **104.24** |
| FFA | 46.29 | 116.68 | 45.90 | 50.26 | 48.24 | **42.73** |
| FFB | 121.75 | 118.36 | 127.43 | 99.70 | 100.76 | **95.85** |
| CFA | 58.54 | 57.91 | 61.37 | **48.03** | 97.81 | 49.26 |
| CFB | 144.72 | 143.76 | 147.69 | 134.42 | 132.53 | **124.02** |

**Table 2 SSIM metrics (higher is better)**

| Dataset | InversionNet | VelocityGAN | Auto-Linear | Latent U-Net | Invertible X-Net | Rectified-flow |
|---------|--------------|-------------|-------------|--------------|------------------|----------------|
| FVA | 0.982 | 0.985 | 0.988 | **0.997** | 0.991 | 0.992 |
| FVB | 0.946 | 0.931 | 0.904 | 0.981 | 0.977 | **0.989** |
| CVA | 0.801 | 0.782 | 0.806 | **0.927** | 0.914 | 0.920 |
| CVB | 0.672 | 0.683 | 0.617 | 0.815 | 0.807 | **0.851** |
| FFA | 0.977 | 0.927 | 0.970 | **0.991** | 0.983 | 0.988 |
| FFB | 0.720 | 0.728 | 0.686 | 0.851 | 0.852 | **0.894** |
| CFA | 0.956 | 0.921 | 0.942 | 0.980 | 0.931 | **0.991** |
| CFB | 0.602 | 0.613 | 0.569 | 0.692 | 0.709 | **0.778** |

## 3.2.2 Rectified Flow: Efficient Sampling

DDPM(Denoising Diffusion Probabilistic Models): execute iterative denoising through a Markov chain, necessitating complete traversal of the forward noising trajectory. Formally, the forward corruption is modeled as a T -step Markov chain; the reverse process consequently mandates strict conditional sampling dependent on the preceding state distribution $p_\theta(x_t-1|x_t)$ , inherently precluding the omission of intermediate states. Standard configurations prescribe $T = 1000$ steps, with each step incurring one neural function evaluation (NFE), aggregating to a total cost of 1000 NFEs. As discretization errors associated with the underlying Stochastic Differential Equation (SDE) accumulate exponentially with increasing step size, the sampling trajectory length remains fundamentally incompressible.

DDIM (Denoising Diffusion Implicit Models): This approach achieves deterministic sampling through a non-Markovian diffusion process, enabling larger step sizes to skip intermediate states. Its theoretical foundation lies in constructing a class of non-Markovian forward processes; by adjusting the variance schedule, the reverse process degenerates into a Probability Flow ODE (PF-ODE), thereby supporting leapfrog sampling. While the sampling steps can be reduced to 200, each step still requires one Neural Function Evaluation (NFE), with total computational cost decreasing linearly with step count. However, large step sizes lead to accumulated truncation errors, causing significant degradation in generation quality below 100 steps.

DPM-Solver: This constitutes a high-order numerical solver for diffusion ODEs, employing a Predictor-Corrector scheme. Leveraging the semi-linear structure of diffusion ODEs, it constructs high-order polynomial approximations via exponential integrators. The predictor phase estimates subsequent states using multi-step methods, while the corrector phase utilizes implicit formulations to rectify local truncation errors. Although capable of reducing sampling steps (e.g., to 50 steps), each iteration requires 2 NFEs (one for prediction plus one for correction). When employing third- or fourth-order schemes, the step count can be further compressed to 20, though the NFEs per step increase correspondingly, rendering total computational cost comparable to second-order alternatives.

Rectified Flow: By learning rectified transport paths that directly map noise to data, this framework theoretically enables single-step generation. Grounded in the Flow Matching paradigm, it directly optimizes the velocity field vt such that the transport trajectory becomes a straight line connecting the noise and data distributions (Euler discretization error vanishes), effectively transforming curved ODEs into straight-line ODEs. In practice, merely 10 or fewer steps suffice to obtain high-quality results, with only 1 NFE required per step, yielding minimal total computational cost. Under the standard 5-step configuration, it achieves Fréchet Inception Distance (FID) scores comparable to 200-step DDIM without necessitating additional training.

Evaluation Protocol. We adopt Number of Function Evaluations (NFE) as the canonical metric for computational expenditure, quantifying the effective number of neural network forward invocations. This formulation offers superior rigor to simplistic "sampling step" counts, given that advanced solvers (e.g., DPM-Solver [17,18]) implicitly entail multiple function evaluations per nominal iteration. Crucially, NFE exhibits direct proportionality to aggregate GPU floating-point operations (FLOPs) and empirical inference latency, thereby establishing an equitable, architecture-neutral benchmark for cross-method efficiency comparison—particularly critical when assessing real-time generation capabilities (e.g., streaming seismic visualization, interactive velocity model editing).

To substantiate the sampling efficiency advantages of Rectified Flow, we conduct comparative benchmarking on the CurveFaultB dataset employing NFE (number of function evaluations) as the canonical metric for computational expenditure. As summarized in Table 3, CRF-FWI achieves comparable reconstruction fidelity (SSIM ≥ 0.76) to DDPM with merely 4 NFEs—contrasted with DDPM's requirement of 1000 NFEs (1000 steps × 1 evaluation per step).

In contrast, although DPM-Solver (2nd order) reduces the sampling steps to 50, its actual NFE cost amounts to 100 due to the predictor-corrector format requiring two network evaluations per step, yielding merely a 10× acceleration. Similarly, DDIM requires 200 NFEs (200 steps × 1 NFE/step) to achieve comparable target quality. These results quantitatively verify the efficacy of the Rectified Flow's straightened transport trajectory—achieving the minimal per-step computational cost while simultaneously requiring the fewest convergence steps—thereby establishing the necessary theoretical and practical foundation for real-time, high-resolution seismic inversion applications.

**Table 3. NFE Computational Cost**

| NFE | Rectified Flow | DDIM | DDPM | DPM-Solver | Best |
|------|------|------|------|------|------|
| 4 | **0.778** | 0.278 | 0.078 | 0.578 | Rectified Flow |
| 20 | **0.775** | 0.413 | 0.125 | 0.632 | Rectified Flow |
| 50 | 0.776 | 0.476 | 0.176 | **0.779** | DPM-Solver |
| 100 | 0.769 | 0.669 | 0.369 | **0.781** | DPM-Solver |
| 200 | 0.768 | 0.758 | 0.568 | **0.776** | DPM-Solver |
| 500 | 0.772 | **0.775** | 0.732 | 0.769 | DDIM |
| 1000 | 0.764 | 0.769 | **0.774** | 0.763 | DDPM |

## 3.3 Ablation Study

### 3.3.1 Ablation Study

To rigorously assess the contribution of individual architectural components to model performance, we conduct systematic ablation experiments on the CurveFaultB datasetencompassing complex dipping fault structures that impose stringent demands on fine-grained conditional information propagation, thereby enabling rigorous evaluation of how distinct injection strategies affect geological boundary preservation. The experimental design tests the central hypotheses articulated in Section 2.3: validating the efficacy of layer-wise injection in alleviating information bottlenecks compared to conventional early-concatenation approaches, and demonstrating the superiority of direct MLP-driven feature interaction over statistical modulation mechanisms (e.g., AdaIN/AdaGN) in preserving seismically consistent physical details.

**Table 4. Conditional Injection and Fusion Mechanisms**

| early injection | Layer-wise Injection | AdaIN | AdaGN | MLP | SSIM | MAE |
|:---:|:---:|:---:|:---:|:---:|:---:|:---:|
| ✓ | | | | ✓ | 0.668 | 131.82 |
| | ✓ | ✓ | | | 0.658 | 135.98 |
| | ✓ | | ✓ | | 0.674 | 132.35 |
| | ✓ | | | ✓ | 0.778 | **124.02** |

The ablation results in Table 4 quantitatively validate the decisive impact of conditional injection strategies on inversion accuracy, revealing two prominent statistical patterns: First, under identical fusion mechanisms (MLP), layer-wise injection (SSIM 0.778, MAE 124.02) significantly outperforms first-layer-only injection (SSIM 0.668, MAE 131.82), achieving a performance gain of 16.5%. This directly corroborates the existence of the "information bottleneck" problem articulated in Section 2.3—when conditional features are concatenated exclusively at the U-Net input layer, high-frequency geological details undergo progressive smoothing and dilution by expanding convolutional receptive fields during hierarchical downsampling, leaving the deep generative process deficient in fine-scale constraints for thin-layer identification and fault delineation. In contrast, the layer-wise modulation-aligned mechanism maintains explicit gradient pathways for conditional signals at each resolution hierarchy, ensuring the complete preservation of multi-scale geological constraints ranging from shallow macro-stratification to deep fine-scale structural features.

### 3.3.2 Ablation Study of Seismic Encoders

To rigorously validate the efficacy and necessity of each constituent component within the seismic encoder architecture delineated in Section 2.2, we conducted systematic ablation studies on the Curve Fault_B dataset. This dataset encompasses intricate dipping fault structures accompanied by pronounced lateral velocity variations, imposing stringent requirements upon the completeness of spatiotemporal wavefield feature extraction. The experiments were designed to quantitatively disentangle the individual contributions and synergistic effects of modules within the progressive three-stage architecture: the temporal convolutional block, which extracts wavefield dynamic characteristics; the spatial convolutional block, which addresses receiver illumination irregularities; and the source aggregation module, which models multi-shot geometric complementarity. Specifically, we constructed eight comparative configurations: a baseline group (three linear layers), single-module groups (temporal convolutional block only, spatial convolutional block only, source aggregation only), dual-module combination groups (temporal-spatial combination, temporal-source combination, spatial-source combination), and the complete group (three-stage cascade). We systematically evaluated the marginal contributions of each module to inversion accuracy (MAE) and structural fidelity (SSIM). Table 5 presents the comparative performance metrics across different encoder configurations.

#### Table 5: Seismic Encoding

| Temporal Convolutional Block | Spatial Convolutional Block | Shot Aggregation | SSIM | MAE |
|:---:|:---:|:---:|:---:|:---:|
| | | | 0.562 | 158.98 |
| ✓ | | | 0.698 | 134.74 |
| | ✓ | | 0.603 | 145.76 |
| | | ✓ | 0.592 | 153.34 |
| ✓ | ✓ | | 0.747 | 137.46 |
| ✓ | | ✓ | 0.734 | 128.99 |
| | ✓ | ✓ | 0.656 | 136.13 |
| ✓ | ✓ | ✓ | **0.778** | **124.02** |

Ablation experiments systematically validate the necessity of the three-stage architecture delineated in Section 2.3. Isolated temporal convolution (Stage 1) yields blurred delineation of geological bou ndaries (SSIM: 0.698) due to deficient spatial structuring capabilities, whereas exclusive reliance on spatial convolution (Stage 2) attains inferior performance (SSIM: 0.603) by neglecting wavefield temporal evolution—consequently failing to discriminate between first arrivals and reflected phases. These observations corroborate that the intrinsically coupled spatiotemporal physics of seismic wave propagation necessitates simultaneous modeling. Upon establishment of comprehensive spatiotemporal feature extraction, source fusion (Stage 3)—realizing multi-shot nonlinear aggregation through 1×1 convolutions—manifests its incremental value: operating in isolation, source fusion achieves the poorest reconstruction fidelity (SSIM: 0.592) owing to the absence of spatiotemporal preprocessing, yet when cascaded with the preceding stages, attains optimal performance (SSIM: 0.778; MAE: 108 m/s). The progressive three-stage processing paradigm—employing learnable temporal pooling to capture wavefield dynamics, followed by spatial attention mechanisms for adaptive weighting of critical receiver positions, and culminating in convolutional nonlinear fusion of multi-source illumination information—effectively preserves multiscale geological signatures spanning from wavefield temporal evolution characteristics to multi-offset spatial geometries.

## 3.3 Feature Visualization of the Seismic Encoder

To systematically validate the physical interpretability of individual seismic encoder components,

comprehensive visual analyses were conducted on the trained architecture using representative test samples. Herein, we probe the learned representations across three complementary dimensions: (i) feature embeddings captured by temporal convolutional modules, (ii) spatial attention allocation patterns, and (iii) channel-wise attention weight distributions.

### 3.3.1 Temporal Convolutional Module Analysis

The learnable temporal convolutional layers (32 filters) in Stage 1 manifest pronounced time-frequency specialization. Analyzing the central shot gather, normalized energy distribution maps of filter responses reveal that specific high-activation channels (e.g., channels 17 and 28, accounting for 16.7% and 16.4% of the energy budget, respectively) peak within the first-arrival window, effectively decoupling raw high-frequency oscillatory signals into smooth energy envelope features; conversely, remaining channels exhibit preferential sensitivity to reflected phases and coda windows. These observations demonstrate that the learnable temporal convolution mechanism spontaneously extracts physically interpretable time windows, enabling adaptive capture and representation of wavefield dynamic characteristics.

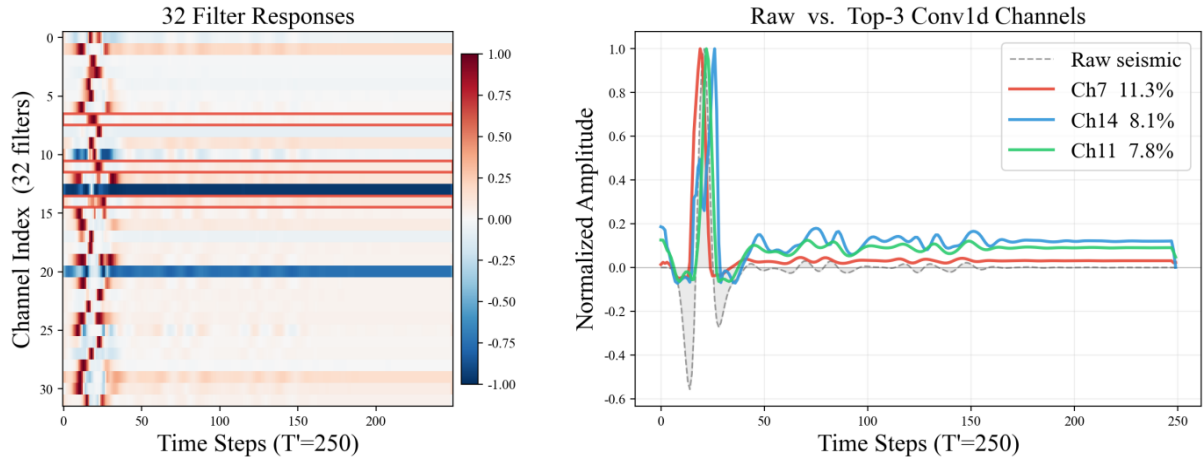

**Figure 8. Visualization of Stage 1 temporal convolutional feature responses.**

The left panel presents a heatmap of normalized responses for the 32 one-dimensional temporal convolutional kernels, wherein red bounding boxes highlight the three channels with highest energy contribution (Top-3). The spectral map reveals that distinct filters have learned differentiated time-frequency selective patterns targeting first arrivals (early time windows) versus reflected phases (late time w indows). The right panel contrasts the raw seismic trace (gray dashed filled curve) against the output features from the Top-3 channels, demonstrating that the Conv1d layer effectively decouples raw high-frequency oscillatory signals into smooth energy envelopes; specifically, channel 17 (16.7%) and channel 28 (16.4%) precisely capture the principal energy distribution of the first-arrival wavefield. These results validate that the learnable temporal downsampling mechanism enables adaptive identification and representation of physically interpretable time windows.

### 3.3.2 Spatial Attention Analysis

Spatial attention masks precisely delineate the hyperbolic moveout patterns characteristic of first-arrival traveltimes. High-response regions corresponding to individual shot gathers align with theoretical traveltime trajectories, accurately tracking the hyperbolic curvature as receiver offset increases—demonstrating that the model spontaneously acquires the kinematic characteristics of seismic wave propagation without explicit physical constraints. Furthermore, attention masks for near- and far-offset sources exhibit mirror symmetry along the receiver axis, consistent with the complementary illumination geometries inherent to field acquisition systems.

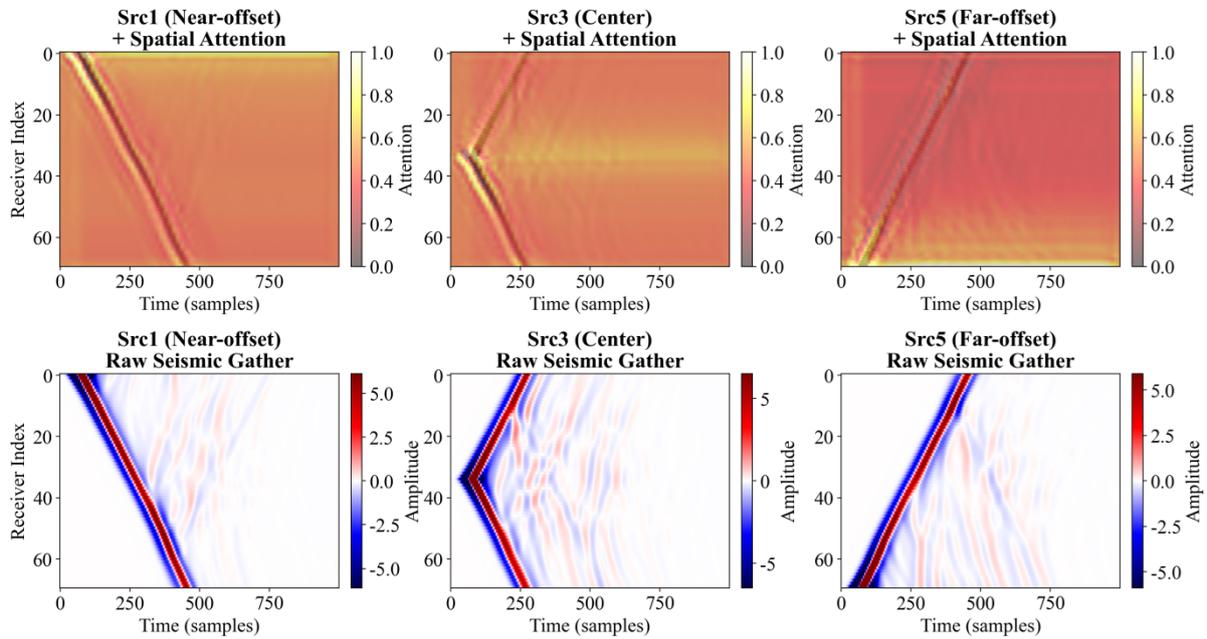

**Figure 9. Comparative visualization of spatial attention (SA).**

The upper panels display the overlay visualization of spatial attention weights superimposed on raw seismic data (from left to right: near-offset, center, and far-offset positions). Regions of high attention activation (warm colors) precisely track the hyperbolic traveltime trajectories of first-arrival waves, while the attention distributions across different shots exhibit mirror-symmetric patterns along the receiver axis consistent with the acquisition geometry. The lower panels show the corresponding raw seismic gathers; comparison indicates that the spatial attention mechanism, without explicit physical constraints, autonomously learns an adaptive receiver-position weighting scheme based on first-arrival traveltimes, effectively highlighting wavefield regions that provide critical constraints for velocity model building.

### 3.3.3 Channel-Wise Attention Analysis

To systematically evaluate the quantitative contribution of Channel Attention (CA) mechanisms to the inversion fidelity, this study designs a suite of controlled perturbation experiments contrasting three distinct weight configurations: (i) preserving the original adaptive CA weights as learned by the network; (ii) retaining solely the top 10% of channels ranked by L2 energy while aggressively masking the remaining 90% to interrogate the information b ottleneck; and (iii) adopting a uniform weighting strategy, effectively equivalent to CA ablation. Through conjoint analysis of the spatial distribution of L2 energy within feature layers and the end-to-end inversion fidelity (quantified via SSIM), we systematically assess the impact of these configurations on both feature representational capacity and inversion accuracy.

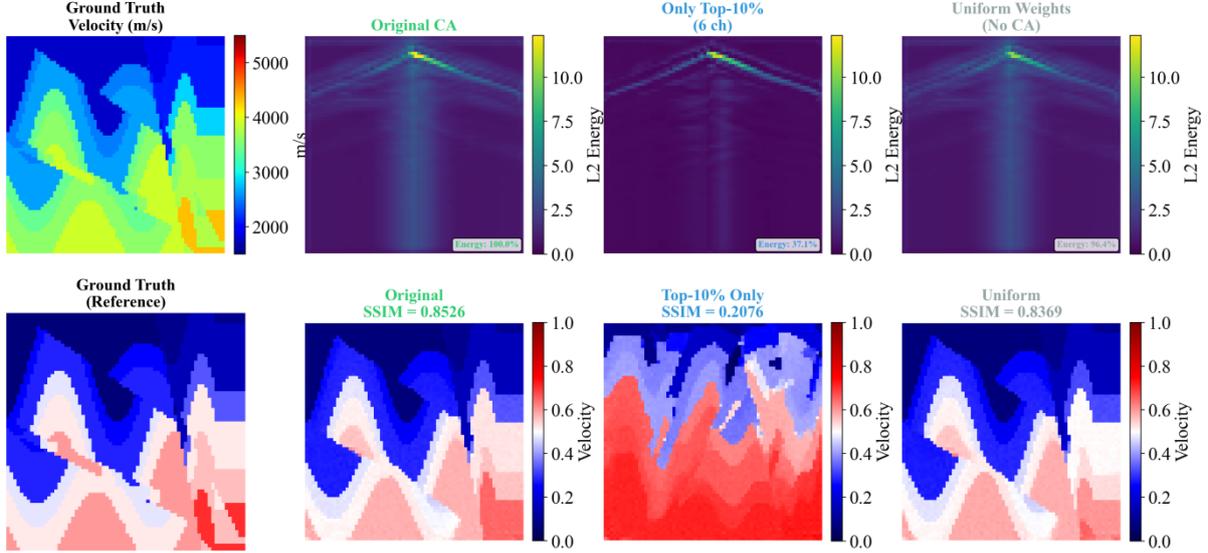

Figure 10. Analysis of channel attention (CA) weights and comparison of feature maps.

Experimental results demonstrate that the channel attention mechanism achieves optimal feature representation and inversion performance through the synergistic utilization of high-, medium-, and low-response channels, rather than relying on individual highly-activated "lone wolf" channels. The aggressive truncation experiment (Top-10%) resulting in catastrophic performance degradation (SSIM decreased by 75.7% relative to baseline) quantitatively attests to the critical role of medium- and low-weight channels in encoding diffractions and stratigraphic structural details. Conversely, the accuracy loss incurred by the uniform weighting strategy (SSIM decreased by 1.8%) substantiates the necessity of a daptive weighting for maintaining feature selectivity. Consequently, the CA mechanism essentially functions as a collaborative perceptual bandwidth allocation strategy: it enhances sensitivity to critical wavefield regions through differentiated weight assignment, while simultaneously preserving subtle structural information latent within weakly-responding channels, thereby guaranteeing globally optimal performance of the end-to-end inversion network.

## 4 Zero-Shot Using the Marmousi Benchmark

Although conventional Full Waveform Inversion (FWI) is grounded in rigorous geophysical theory, its inversion fidelity remains critically dependent upon the initial velocity model. Suboptimal starting models frequently induce cycle-skipping artifacts and convergence to local minima, while substantially escalating computational costs. While CRF-FWI has demonstrated efficacy on the OpenFWI benchmark dataset, its zero-shot generalization to the Marmousi model—despite outperforming alternative deep generative approaches—still exhibits limited accuracy. To address this limitation, we propose leveraging the computationally efficient, zero-shot predictions of CRF-FWI as the initial model for conventional FWI, thereby circumventing the initialization sensitivity inherent to traditional waveform inversion. This section validates the efficacy of our CRF-FWI methodology for zero-shot construction of FWI initial velocity models using the standard Marmousi geological benchmark.

## 4.1 Experimental Setup

The Marmousi model constitutes a quasi-canonical two-dimensional synthetic benchmark in exploration geophysics, encompassing intricate structural elements including steeply dipping faults, high-velocity salt domes, and thinly interbedded strata. Its geological complexity markedly supersedes that of the OpenFWI StyleA dataset employed for training, thereby establishing this framework as an industry-standard stress test for evaluating the generalization capabilities of seismic waveform inversion methodologies.

This study strictly enforces zero-shot testing protocols: the model undergoes training solely on the OpenFWI StyleA dataset, with strict prohibition of exposure to Marmousi-related data or weight fine-tuning, directly generating initial velocity models from Marmousi forward-modeled seismic records.

## 4.2 Experimental Results and Analysis

Comparative analyses between our framework and existing methods are presented.

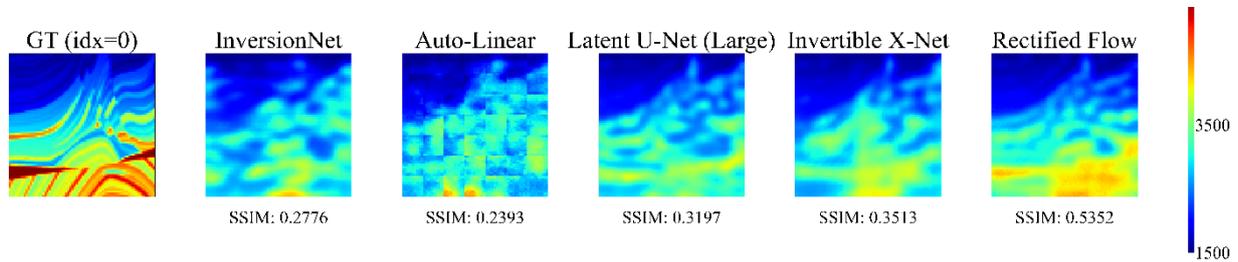

**Figure 11. Cross-domain generalization comparison on the Marmousi model.**

Figure 11 presents the inversion results of our proposed framework alongside baseline methodologies applied to the Marmousi benchmark. Despite training exclusively on the OpenFWI synthetic corpus, the model exhibits robust out-of-distribution generalization, achieving faithful reconstructions on this structurally complex, zero-shot geological scenario. Relative to competing approaches, our method demonstrates superior delineation accuracy for fault geometries and salt-dome boundaries, coupled with enhanced velocity fidelity that closely approximates the true subsurface model.

The observed zero-shot generalization performance stems from three architectural insights: (1) the Rectified Flow framework learns the underlying structural priors of geophysical data distributions rather than engaging in dataset-specific memorization; (2) the seismic encoder extracts transferable wavefield representations that exhibit robust cross-domain applicability; and (3) the hierarchical conditioning strategy enhances the model's adaptability to multi-scale geological features across varying spatial resolutions.

## 5 Discussion and Conclusions

This study presents an end-to-end rapid seismic waveform inversion methodology predicated on the Conditional Rectified Flow framework. Through architectural innovations encompassing dedicated seismic encoder networks and layer-wise conditional injection mechanisms, the proposed approach achieves high-fidelity subsurface reconstruction while maintaining exceptional computational efficiency suitable for operational deployment. The principal conclusions are summarized as follows:

- The Rectified Flow framework achieves two-order-of-magnitude acceleration (5 vs. 500 steps) without compromising geophysical fidelity;
- Dedicated seismic encoders with layer-wise conditional injection enhance adaptability to complex geological structures;
- Robust zero-shot generalization enables rapid generation of high-fidelity velocity models, providing quality initializations for conventional FWI and mitigating its stringent dependence on accurate starting models, with substantial industrial deployment potential.

Future research will be directed toward the following avenues: (1) scaling the framework to three-dimensional seismic inversion scenarios; (2) incorporating physics-informed constraints to further enhance the geological plausibility of recovered models; and (3) investigating few-shot learning paradigms to mitigate reliance on extensively annotated training datasets.